\documentclass[a4paper, 10 pt, conference]{IEEEtran}  

\IEEEoverridecommandlockouts                              

\usepackage{cite}
\usepackage{amsmath,amssymb,amsfonts}
\usepackage{algorithmic}
\usepackage{graphicx}
\usepackage{textcomp}
\usepackage{xcolor}
\usepackage{enumerate}
\usepackage{algorithm}
\usepackage{algorithmic}
\usepackage{multirow}
\usepackage{booktabs}
\usepackage{amsfonts}
\usepackage{microtype} 
\usepackage{doi}
\title{\LARGE \bf
Continual Learning via Manifold Expansion Replay
}

\author{
    \IEEEauthorblockN{Zihao Xu$^1$, Xuan Tang$^2$, Yufei Shi$^3$, Jianfeng Zhang$^1$, Jian Yang$^{4}$, Mingsong Chen$^1$, Xian Wei$^{1*}$}\\
    \small
    \IEEEauthorblockA{$^1$ Software Engineering Institute, East China Normal University, Shanghai, China}\\
    \IEEEauthorblockA{$^2$ School of Communication \& Electronic Engineering, East China Normal University, Shanghai, China}\\
    \IEEEauthorblockA{$^3$ Department of Medical Informatics and Zhongshan School of Medicine, Sun Yat-sen University, Guangzhou, China}\\
    \IEEEauthorblockA{$^4$ School of Geospatial Information, Information Engineering University, Zhengzhou, China}\\
    \IEEEauthorblockA{
    xian.wei@tum.de
    }\\
    \thanks{$^*$ Corresponding author.}
}

\begin{document}

\maketitle
\thispagestyle{empty}
\pagestyle{empty}

\begin{abstract}
In continual learning, the learner learns multiple tasks in sequence, with data being acquired only once for each task. Catastrophic forgetting is a major challenge to continual learning. To reduce forgetting, some existing rehearsal-based methods use episodic memory to replay samples of previous tasks. However, in the process of knowledge integration when learning a new task, this strategy also suffers from catastrophic forgetting due to an imbalance between old and new knowledge. To address this problem, we propose a novel replay strategy called Manifold Expansion Replay (MaER). We argue that expanding the implicit manifold of the knowledge representation in the episodic memory helps to improve the robustness and expressiveness of the model. To this end, we propose a greedy strategy to keep increasing the diameter of the implicit manifold represented by the knowledge in the buffer during memory management. In addition, we introduce Wasserstein distance instead of cross entropy as distillation loss to preserve previous knowledge. With extensive experimental validation on MNIST, CIFAR10, CIFAR100, and TinyImageNet, we show that the proposed method significantly improves the accuracy in continual learning setup, outperforming the state of the arts.

\end{abstract}

\begin{IEEEkeywords}
continual learning, catastrophic forgetting, manifold diameter, Wasserstein distance
\end{IEEEkeywords}

\section{Introduction}
Continual learning, also known as incremental learning \cite{aljundi2018memory,chaudhry2018riemannian,aljundi2017expert}, refers to the process of sequentially learning multiple tasks without forgetting previous knowledge. In this setting, catastrophic forgetting \cite{french1999catastrophic,ratcliff1990connectionist} is a major challenge for continual learning, where previously learned knowledge is lost when learning new tasks.

Continual learning has recently gained increasing attention in the field of artificial intelligence. Various strategies have been proposed to overcome catastrophic forgetting \cite{de2019continual},  including rehearsal-based strategies \cite{rolnick2019experience,ayub2020storing,aljundi2019gradient}, regularization-based strategies \cite{zenke2017continual,kirkpatrick2017overcoming}, and parameter isolation-based strategies \cite{rusu2016progressive, li2019learn}. These strategies are mutually orthogonal and can be combined in a specific scenario. Among these strategies, rehearsal-based methods have proven to be a simple yet effective approach that uses episodic memory to replay training samples.  Despite its encouraging success, there are still challenges that need to be addressed, including the issue of overfitting and biased knowledge representation due to knowledge imbalance in episodic memory. A naive but effective solution is increasing the memory size when new samples come. However, this approach increases the memory requirement and violates the setting of limited memory resource requirements in continual learning.

To address this issue, we propose a novel replay strategy called Manifold Expansion Replay (MaER). MaER investigates two factors to improve neural network performance in continual learning settings. Firstly, MaER views the process of continual learning as a fusion of implicit manifolds represented by knowledge. When the diameters of manifolds are imbalanced, the larger one will receive more bias while the smaller one will experience forgetting. Inspired by this, MaER adopts a greedy sampling strategy to manage memory, helping the neural network to learn unbiased presentation of all data.  Secondly, MaER introduces the Wasserstein distance as distillation loss. The Wasserstein distance between two distributions is defined as the minimum cost required to match one distribution with another. Unlike traditional distance metrics such as Euclidean distance, Wasserstein distance considers the underlying structure of the compared distributions, which can help the neural network better fuse knowledge manifolds. 

We mainly evaluate MaER on Permuted MNIST, Rotated MNIST, Split CIFAR10, Split CIFAR100, and Split TinyImageNet datasets. The extensive ablation studies and experimental results show that MaER gains significant performance improvement, outperforming state-of-the-art in accuracy.

Our contributions are summarized as follows:
\begin{itemize}
\item We propose a greedy sampling strategy to balance knowledge by expanding the diameter of the knowledge manifold in episodic memory.
\item We propose to distill knowledge using Wasserstein distance, which helps neural networks effectively fuse knowledge in continual learning.

\end{itemize}

\section{Related Work}
Here we briefly review previous research works. Existing works can be divided into three categories, i.e., rehearsal-based, regularization-based, and parameter isolation-based.

\paragraph{Rehearsal-based Strategy} 
The rehearsal-based strategy can be viewed as a review strategy that uses a capacity-limited buffer called episodic memory to replay a portion of the samples from the previous task at each training session. Despite its simplicity, this rehearsal strategy has been shown to be effective and work well when large memories are available. Typical approaches include Incremental Classifier and Representation Learning (iCaRL) \cite{rebuffi2017icarl}, Experience Replay (ER) \cite{rolnick2019experience}, Selective Experience Replay (SER) \cite{isele2018selective}, Continual Prototype Evolution (CoPE) \cite{de2021continual} and Tiny Experience Replay (TEM) \cite{chaudhry2019continual}.

\paragraph{Regularization-based Strategy} 
In contrast to the rehearsal strategy, the regularization strategy adopts a more strict continual learning setup. This strategy aims to reduce forgetting without accessing prior task data. Existing methods can be further divided into two categories, i.e., Data-focused and prior-focused \cite{de2021continual}. The main idea of data-focused methods is to transfer knowledge from a teacher model to a student model using the knowledge distillation technique, where the teacher model has been trained on previous tasks. The idea of using knowledge distillation to improve performance on new tasks in continual learning was first proposed by \cite{silver2002task}. 
Subsequent research has proven that knowledge distillation can also reduce forgetting. Typical methods include Learning without Forgetting (LwF) \cite{li2017learning}, Learning from Less (LFL) \cite{jung2016less}, and Dark Knowledge distillation with Memory Consolidation (DMC) \cite{zhang2020class}.
The basic idea of prior-focused methods is to estimate the importance of parameters to previous tasks and then penalize changes to important parameters during training to prevent catastrophic forgetting. This strategy has been shown to be effective. Typical methods include Elastic Weight Consolidation (EWC) \cite{kirkpatrick2017overcoming}, Variational Continual Learning (VCL) \cite{nguyen2017variational}, Incremental Moment Matching (IMM) \cite{lee2017overcoming}, Synaptic Intelligence (SI) \cite{zenke2017continual}, and Riemannian Walk (RW) \cite{chaudhry2018riemannian}.

\paragraph{Paramters Isolation-based Strategy} This strategy overcomes catastrophic forgetting by selecting parameters from a fixed network or dynamically modifying the network structure. Typical methods, selecting a subnetwork for each task, include Hard Attention (HAT) \cite{serra2018overcoming}, PackNet \cite{mallya2018packnet}, and PathNet \cite{fernando2017pathnet}. HAT uses a hard attention mask to selectively prune network parameters, retaining important features relevant to the current task and reducing forgetting. PackNet uses the network pruning technique to make the network adapt to multiple tasks. PathNet divides the network into sub-networks, with each sub-network responsible for each task.  Dynamically expanding network structure and establishing new neural connections for new tasks has also been proven to be an effective strategy. Typical methods include Progressive Neural Networks (PNN) \cite{rusu2016progressive}, Deep Adaptive Network (DAN) \cite{rosenfeld2018incremental}, and Reinforced Continual Learning (RCL) \cite{xu2018reinforced}. Both PNN and DAN adopt a hierarchical structure, where new network layers are established for new tasks, and each layer is responsible for a specific task. RCL, on the other hand, employs reinforcement learning techniques to adjust the network's learning strategy by rewarding and punishing its learning process.
%
%
\section{Continual Learning Setup}
In continual learning, a learner needs to sequentially learn $T$ tasks 
$\{(\mathcal{X}_1,\mathcal{Y}_1,...,(\mathcal{X}_T,\mathcal{Y}_T)\}$.
$(\mathcal{X}_t,\mathcal{Y}_t)$ represents a dataset $\mathcal{D}_t = \{ (x_t^1,y_t^1),...,(x_t^{n_t},y_t^{n_t})\}$ from task $t$, randomly sampled from an unknown distribution $\mathcal{P}_t$, where $x_t$ represents the sample, $y_t$ represents the corresponding ground truth label, and $n_t$ represents the number of samples in the dataset. We assume that the learner can use a capacity-limited buffer $\mathcal{M}$ to store a small number of samples during learning. 
Our goal is to train a predictor $f = (w \circ \Phi): \mathcal{X} \rightarrow \mathcal{Y}$, composed of a feature extractor $\Phi$ and a classifier $w$, that minimizes the risk over all the data it has seen while only having access to a limited number of samples from previous tasks that are stored in $\mathcal{M}$:
\begin{equation}
    \frac{1}{T} \sum_{t=1}^{T} \mathbb{E}_{(x, y) \sim P_{t}}[\ell(f(x; \theta), y)],
\end{equation}
where $\theta$ denotes the parameters of model and $\ell$ denotes the loss function.

\textbf{Evaluation Metrics.} Following \cite{chaudhry2018riemannian,chaudhry2019continual,lopez2017gradient}, we use average classification accuracy (ACC) and backward transfer (BWT) to evaluate performance. Formally, ACC is defined as:
\begin{equation}
    \text { ACC }=\frac{1}{T} \sum_{j=1}^{T} a_{T, j},
\end{equation}
where $a_{i,j}$ denotes the test accuracy of the model on task $j$ after learning task $i$. BWT is defined as the average change in accuracy of old tasks after learning a new task:

\begin{equation}
    \text {BWT}=\frac{1}{T-1} \sum_{j=1}^{T-1} \max _{l \in\{1, \ldots, T-1\}}\left(a_{l, j}-a_{T, j}\right).
\end{equation}
A positive value of BWT means that learning a new task benefits the old tasks, while a negative value indicates that learning a new task interferes with the old tasks.

\begin{table*}[h!]
    \centering
    \caption{\label{main_table} Classification results for Split CIFAR10, Split TinyImageNet, Permuted MNIST and Rotated MNIST. }    
    \begin{tabular*}{\hsize}{@{}@{\extracolsep{\fill}}cl|cccc@{}}
    \toprule  
    \textbf{Buffer} &  \textbf{Method} & \textbf{Split CIFAR10} & \textbf{Split TinyImageNet} & \textbf{P-MNIST} & \textbf{R-MNIST} \\
    \midrule
    \multirow{2}{*}{-} & JOINT & $98.31 \pm 0.12$ & $82.04 \pm 0.10$ & $94.33 \pm 0.17$ & $95.76 \pm 0.04$ \\
                          & SGD   & $61.02 \pm 3.33$ & $18.31 \pm 0.68$ & $40.70 \pm 2.33$ & $6.77 \pm 8.53$ \\
    \cmidrule(){1-6}
    \multirow{4}{*}{-} & oEWC & $68.29 \pm 3.92 $ & $19.20 \pm 0.31$ & $\textbf{75.79} \pm \textbf{2.25}$ & $\textbf{77.35} \pm \textbf{5.77}$ \\
                       & SI & $68.05 \pm 5.91 $ & $36.32 \pm 0.13$ & $65.86 \pm 1.57$ & $71.91 \pm 5.83$ \\
                       & LwF & $63.29 \pm 2.35 $ & $19.20 \pm 0.31$ & - & - \\
                       & PNN & $\textbf{95.13} \pm \textbf{0.72} $ & $\textbf{67.84} \pm \textbf{0.29}$ & - & - \\
    \cmidrule(){1-6}
    \multirow{7}{*}{200} & ER & $91.19 \pm 0.94$ & $ 38.17 \pm 2.00$ & $72.37 \pm 0.87$ & $85.01 \pm 1.90$ \\
                         & A-GEM & $83.88 \pm 1.49$ & $22.77 \pm 0.03$ & $66.42 \pm 4.00$ & $81.91 \pm 0.76$ \\
                         & iCaRL & $88.99 \pm 2.13$ & $28.19 \pm 1.47$ & - & - \\
                         & HAL   & $82.51 \pm 3.20$ & - & $74.16 \pm 1.65$ & $84.02 \pm 0.98$ \\
                         & DER   & $91.40 \pm 0.92$ & $40.22 \pm 0.67$ & $81.74 \pm 1.07$ & $90.04 \pm 2.61$ \\
                         & SNCL  & $\textbf{92.91} \pm \textbf{0.81}$ & $43.01 \pm 1.67$ & $86.23 \pm 0.20$ & $91.54 \pm 2.58$ \\
                         & \textbf{MaER (ours)}  & $92.56 \pm 0.49$ & $\textbf{46.34} \pm \textbf{0.79}$ & $\textbf{90.04} \pm \textbf{0.28}$ & $\textbf{91.88} \pm \textbf{1.96}$ \\
    \cmidrule(){1-6}
    \multirow{7}{*}{500} & ER & $93.61 \pm 0.27$ & $ 48.64 \pm 0.46$ & $80.60 \pm 0.86$ & $88.91 \pm 1.44$ \\
                         & A-GEM & $89.48 \pm 1.45$ & $25.33 \pm 0.49$ & $67.56 \pm 1.28$ & $80.31 \pm 6.29$ \\
                         & iCaRL & $88.22 \pm 2.62$ & $31.15 \pm 3.27$ & - & - \\
                         & HAL   & $84.54 \pm 2.36$ & - & $80.13 \pm 0.49$ & $85.00 \pm 0.96$ \\
                         & DER   & $93.40 \pm 0.39$ & $51.78 \pm 0.88$ & $87.29 \pm 0.46$ & $92.24 \pm 1.12$ \\
                         & SNCL  & $\textbf{94.02} \pm \textbf{0.43}$ & $52.85 \pm 0.67$ & $88.53 \pm 0.41$ & $\textbf{93.05} \pm \textbf{1.02}$ \\
                         & \textbf{MaER (ours)}  & $93.29 \pm 0.42$  & $\textbf{54.65} \pm \textbf{0.77}$ & $\textbf{92.34} \pm \textbf{0.54}$ & $92.55 \pm 1.08$\\
    \bottomrule                 
    \end{tabular*}
\end{table*}

\begin{table*}[!h]
\caption{\label{table_2}Average accuracy (ACC) and forgetting (BWT) results on Split CIFAR100.}
\begin{tabular*}{\hsize}{@{}@{\extracolsep{\fill}}lcccccc@{}}
\toprule  
\textbf{Method}                       & \multicolumn{6}{c}{\textbf{Episodic Memory}}                \\ 
\midrule
\multicolumn{1}{c}{\multirow{2}{*}{}} & \multicolumn{3}{c}{ACC}           & \multicolumn{3}{c}{BWT} \\ 
\cmidrule(){2-7}
\multicolumn{1}{c}{}                  & 100 & 300 & 500                   & 100    & 300    & 500   \\ 
\cmidrule(){2-7}
A-GEM                                 &  $54.9 \pm 2.92$   & $56.9 \pm 3.45$    & \multicolumn{1}{c|}{$59.9 \pm 2.64]$} &  $0.14 \pm 0.03$      &  $0.13 \pm 0.03$      &  $0.10 \pm 0.02$     \\
ER                                   &  $49.7 \pm 2.97$   &  $57.7 \pm 2.59$   & \multicolumn{1}{c|}{$60.6 \pm 2.09$} &   $0.19 \pm 0.03$    &  $0.11 \pm 0.01$      &   $0.09 \pm 0.02$    \\
ER-RING                               &  $56.2 \pm 1.93$   &  $60.9 \pm 1.44$   & \multicolumn{1}{c|}{$62.6 \pm 1.77$} &  $0.13 \pm 0.01$      &  $0.09 \pm 0.01$      &  $0.08 \pm 0.02$     \\
ER-RESERVOIR                          &  $53.1 \pm 2.66$   & $59.7 \pm 3.87$    & \multicolumn{1}{c|}{$65.5 \pm 1.99$} &   $0.19 \pm 0.02$     &  $0.12 \pm 0.03$      &  $0.09 \pm 0.02$     \\
\textbf{MaER(ours)}                   &  $\textbf{57.46} \pm \textbf{0.95}$  &  $\textbf{62.61} \pm \textbf{1.59}$   & \multicolumn{1}{c|}{$\textbf{66.4} \pm \textbf{1.56}$} &   $0.19 \pm 0.01$     &    $0.13 \pm 0.01$    &  $0.09 \pm 0.01$     \\ 
\cmidrule(){2-7}
FINETUNE                              &  $40.6 \pm 3.83$   &  -   &         -              &     -   &    -    &   -    \\
EWC                                   &  $41.2 \pm 2.67$   &  -   &        -               &    -    &     -   &    - 
\\
\bottomrule    
\end{tabular*}
\end{table*}
\begin{table}[!h]
\caption{\label{table_3} Ablation study of the different components in proposed method. $\mathcal{L}_{WD}$ denotes the wasserstein distance.}    
\begin{tabular*}{\hsize}{@{}@{\extracolsep{\fill}}lcccc@{}}
\toprule  
\multirow{2}{*}{\textbf{Method}}           & \multirow{2}{*}{\textbf{Dataset}} & \multicolumn{3}{c}{\textbf{\begin{tabular}[c]{@{}c@{}}ACC (\%)\\ Memory size\end{tabular}}} \\ \cmidrule{3-5} 
                                           &                                   & 100                          & 300                          & 500                          \\ \midrule
\multirow{2}{*}{$\mathcal{L}_{CE}$}                        & P-MNIST        & $77.64$   &          $82.62$                    &          $85.38$                    \\
                                           & Split CIFAR100                    &    $56.43$  & $59.64$            &  $64.05$                            \\ \cmidrule{1-5}
\multirow{2}{*}{$\mathcal{L}_{CE} + \mathcal{L}_{WD}$} & P-MNIST                           &             $81.79$                 &      $87.72$                        &        $90.14$                      \\
                                           & Split CIFAR100                    &  $57.49$                            &    $62.46$            &   $66.37$                             \\ \cmidrule{1-5}
\multirow{2}{*}{MaER}                      & P-MNIST                           &   $84.47$                            &      $89.52$         &     $92.07$                        \\
    & Split CIFAR100                           &  $57.99$ &      $63.95$         &     $67.86$                         \\ \bottomrule
\end{tabular*}
\end{table}

\section{Manifold Expansion Replay}
Recent research has shown that the replay strategy is a simple and effective approach to continual learning. However, to develop more robust methods based on the replay strategy, two issues need to be considered: (1) how to replay during the training phase and (2) how to sample and manage memory after each task training. Our approach, MaER, designs strategies for these two stages from a geometric perspective.
\subsection{How to Replay}
In the replay strategy, the number of samples collected for replay is limited. This poses a challenge in recalling the knowledge of the entire task from these samples. If a classification loss is used when training replay samples, the model can only learn to classify these samples rather than previous tasks. As a result, the model may still suffer from catastrophic forgetting as the number of tasks increases. For a specific task, we assume that each sample represents a piece of meta-knowledge. The model learns the entire knowledge manifold by learning from these samples. We now consider the first problem encountered: how to integrate this meta-knowledge when learning new tasks to form a more comprehensive knowledge manifold. To do this, we need to measure the distance between two knowledge manifolds. Intuitively, the distance between the new and old knowledge manifolds should be small because previously acquired meta-knowledge does not change. If we view the knowledge manifold as a distribution of meta-knowledge, one possible choice for a distance metric is KL divergence, which is commonly used to measure the distance between two distributions. However, KL divergence cannot be used as a strict distance function because it is asymmetric and cannot provide distance information when two distributions do not overlap. Our method, MaER, uses Wasserstein distance to measure the distance between two knowledge manifolds.

In mathematics, the Wasserstein distance is a distance function defined between probability distributions on a given metric space $(M,\rho)$, where $\rho(x,y)$ is a distance function for two instances $x$ and $y$ in the set $M$. Formally, the $p$-th Wasserstein distance between two probability measures $\mu$ and $\nu$ on $M$ with $p$-moment is defined as:
\begin{equation}
    W_{p}(\mu, \nu):=\left(\inf _{\gamma \in \Gamma(\mu, \nu)} \int_{M \times M} d(x, y)^{p} \mathrm{~d} \gamma(x, y)\right)^{1 / p},
\end{equation}
where $\Gamma(\mu,\nu)$ represents the set of all coupling of $\mu$ and $\nu$.
Wasserstein distance has some good properties. In contrast to KL divergence, Wasserstein distance is a symmetric metric and provides information even if the distributions do not overlap. 

Now we describe how MaER uses the Wasserstein distance to facilitate meta-knowledge fusion in continual learning. When learning task $i$, we have a teacher model $f_t$ that has learned the previous $i-1$ tasks and a student model $f_s$ responsible for learning task $i$. For a sample $x$, we use $\Phi_s(x;\theta)$ to denote the knowledge representation of $f_s$ for that sample, where $\Phi_s$ denotes the feature extractor for $f_s$.
Our objective is for the student model $f_s$ to effectively learn task $i$. To achieve this, we train $f_s$ on samples from task $i$ using the cross-entropy loss function for classification. Concurrently, it is imperative that $f_s$ retains the knowledge acquired from previous tasks. During training, we replay samples from the memory buffer $\mathcal{M}$, and in addition to learning to classify these samples accurately, we aim to minimize the Wasserstein distance between the knowledge representation learned by $f_s$ and that of the teacher model $f_t$.
Taking these considerations into account, the new loss function is defined as: 
\begin{equation}
    \begin{aligned}
        \mathcal{L} = &\mathbb{E}_{(x, y) \sim D_{t}}[\ell(f_s(x; \theta), y)] \\
        &+ \mathbb{E}_{(x,y) \sim \mathcal{M}}[\ell(f_s(x;\theta),y) + W_p(\Phi_s(x;\theta),\Phi_t(x;\theta))],     
    \end{aligned}
\end{equation}
where $\ell$ is the cross-entropy function in classification tasks. For ease of computation, we use the Wasserstein distance with p = 2 and assume that meta-knowledge is equally important. In MaER, the calculation of $W_2$ can be simplified to the following form: 
\begin{equation}
    W_2(P,Q) = (\frac{1}{n}\sum_i^n d(x, y)^2)^{1/2}, 
\end{equation}
where $d$ is the distance function in metric space $M$, which can be Euclidean distance or geodesic distance.

\begin{algorithm}[!h]
\renewcommand{\algorithmicrequire}{\textbf{Input:}}
\renewcommand{\algorithmicensure}{\textbf{Output:}}
      \caption{Manifold Expansion Sampling}     
      \label{MES}
     \begin{algorithmic} 
        
    \STATE{\bfseries Input:} $\Phi_s, D_t, \mathcal{M}, mem\_size, n$
    \STATE {\bfseries Output:} $\mathcal{M}$
        \STATE $j \leftarrow 0$
        \FOR{ $(x,y)$ in $\mathcal{D}_t$}
            \IF{$|\mathcal{M}| < mem\_size$}
                \STATE $\mathcal{M}$.append($x,y$)
            \ELSE
                \STATE $feature\_\mathcal{M} \leftarrow \Phi_s(\mathcal{M})$
                \STATE $\mathcal{C}, diameter \leftarrow \operatorname{CentroidAndDiameter}(feature\_\mathcal{M})$ 

                \STATE $feature\_x \leftarrow \Phi_s(x)$
                \IF{$\operatorname{distance}(\mathcal{C}, feature\_x) > diameter$}
                    \STATE $i \leftarrow \operatorname{randint}(0, |\mathcal{M}|)$
                    \STATE $\mathcal{M}[i] \leftarrow (x,y)$
                \ELSE
                    \STATE $i \leftarrow \operatorname{randint}(0, n + j)$
                    \IF{$i < mem\_size$}
                        \STATE $\mathcal{M}[i] \leftarrow (x,y)$
                    \ENDIF
                \ENDIF
            \ENDIF
            \STATE $j \leftarrow j + 1$
        \ENDFOR
    \STATE $n \leftarrow n + j$
    \STATE {\bfseries Return} $\mathcal{M}$
  \end{algorithmic} 
\end{algorithm}

\subsection{How to Manage Memory}
We now turn to the second problem: how to sample such that a small number of meta-knowledge can represent the knowledge manifold of the entire task as much as possible. We assume that the meta-knowledge is uniformly distributed over this knowledge manifold. We can only sample a small fraction of meta-knowledge to represent the whole manifold. Intuitively, our sampling method needs to account for two aspects: (1) sample as uniformly as possible to maintain the geometric properties, (2) the sampled meta-knowledge should span the entire knowledge manifold as much as possible, avoiding bias towards new knowledge and preventing forgetting during learning.
To address this problem, MaER employs a greedy strategy that incrementally enlarges the diameter of the manifold during the sampling process. First, the sampling process is stochastic. When a new sample arrives, if $\mathcal{M}$ is already full, then each sample in $\mathcal{M}$ has an equal chance of being replaced. Stochastic sampling preserves the consistency of the data distribution in $\mathcal{M}$ and the original distribution. Second, MaER’s sampling strategy guarantees that samples that can augment the diameter of the manifold are always collected. For other samples, MaER will collect them with a certain probability. The criterion for determining whether a sample can augment the diameter of the manifold is essential for understanding MaER’s sampling strategy. We begin by introducing the concepts of the centroid and diameter of a manifold. The Fréchet mean is a natural generalization of the centroid and can be applied to any manifold. For a metric space $\mathcal{X} = (X,d,\mu)$, its Fréchet mean is defined as:
\begin{equation}
    \underset{x \in X}{\arg \min } \int_{X}d^2(x,y)d\mu(y).    
\end{equation}
With the manifold centroid $\mathcal{C}$, we can estimate the diameter of the manifold in a simple and efficient way. 
We define the diameter as the largest distance from the centroid $\mathcal{C}$ to the sample $x$. Having defined these concepts, we can describe the sampling process for MaER, as shown in Algorithm \ref{MES}.


\section{Experiments}
We apply MaER to different sequential tasks for continual learning and compare it with state-of-the-art replay methods, and then we empirically analyze the proposed algorithm.
\subsection{Experimental Setting}
Here, we begin by describing the continual learning benchmarks, implementation details, and compared methods.

\textbf{Benchmarks.} 
We conducted experiments on several continual learning datasets: Permuted MNIST, Rotated MNIST, Split CIFAR10, Split CIFAR100 and Split TinyImageNet. Permuted MNIST is derived from applying a random permutation to the pixels. Rotated MNIST is derived from rotating the image at a random angle. Split CIFAR10, Split CIFAR100, and Split TinyImageNet are derived from splitting CIFAR10, CIFAR100, and miniImageNet, respectively, such that the classes in the different tasks are disjoint. Split CIFAR10 consists of 5 tasks, while the other datasets consist of 20 tasks each.

\textbf{Implementation details.}
In the network architecture, we utilized a three-layer MLP with 256 neurons in the hidden layer for MNIST and a standard resnet18 for CIFAR10, CIFAR100 and TinyImageNet. We optimize the parameters during training using SGD. The learning rate is set to 0.01 for MNIST and 0.003 for others. The batch size is set to 16 for all experiments. We trained the model for 10 epochs on MNIST, 5 epochs on Split CIFAR100 and 20 epochs on others.

\textbf{Compared methods.}
We compare MaER with several baseline methods. The rehearsal-based baselines include iCaRL \cite{rebuffi2017icarl}, ER \cite{riemer2018learning}, ER-RING \cite{chaudhry2019tiny}, ER-RESERVOIR \cite{chaudhry2019tiny}, A-GEM \cite{chaudhry2018efficient}, HAL \cite{chaudhry2021using}, DER \cite{buzzega2020dark}, and SNCL \cite{yan2022learning}, with DER and SNCL being the strongest baselines. We also compare with other strategy-based methods, including SI \cite{zenke2017continual}, EWC \cite{kirkpatrick2017overcoming}, oEWC \cite{schwarz2018progress}, LwF \cite{li2017learning}, and PNN \cite{rusu2016progressive}. Additionally, we compare with two non-continual learning methods, JOINT and SGD, as upper and lower bounds.

\subsection{Main Results}
Table \ref{main_table} shows the results of our method using a tiny buffer on Split CIFAR10, Split TinyImagNet, Permuted MNIST, and Rotated MNIST. On Split CIFAR10 and Rotated MNIST, our method achieved competitive results, differing from state-of-the-art replay methods by $0.5\% \sim 0.73\%$. On Permuted MNIST and Split TinyImageNet, our method achieves state-of-the-art performance, significantly outperforming baseline methods. On Split TinyImageNet, MaER surpassed SNCL by $1.80\% \sim 3.33\%$ in average accuracy when taking different buffer sizes. On Permuted P-MNIST, this gap was even more pronounced, with our method leading SNCL by $3.81\%$ in average accuracy when taking different buffer sizes.
On Split CIFAR10 and Rotated MNIST, most baselines achieved considerable results. However, on Permuted MNIST and Split TinyImageNet, the gap between different methods became more pronounced. These results indicate that Permuted MNIST and Split TinyImageNet are more challenging when using a tiny buffer. MaER can work well in challenging scenarios and achieves state-of-the-art in average accuracy.
On Split CIFAR100, we also compared MaER with more replay strategies. In Table \ref{table_2}, our method surpassed all baselines when taking different buffer sizes, leading the baseline by $0.9\% \sim 7.64\%$ in average accuracy. As the buffer size increased, MaER was able to perform correspondingly better.
Table \ref{table_3} shows the ablation study results of different components of MaER. Here, $\mathcal{L}_{WD}$. represents the Wasserstein distance loss. Compared to the naive replay, both the replay strategy and memory management in MaER achieve certain performance improvements.

\section{Conclusion}
In this paper, we propose a new replay strategy called MaER. MaER employs knowledge distillation techniques and introduces the Wasserstein distance between the features of the teacher and student models as a distillation loss to integrate old and new knowledge better. Intuitively, when integrating two imbalanced knowledge manifolds, the larger manifold will receive more bias, leading to catastrophic forgetting. MaER addresses this issue through manifold expansion sampling. Samples that can expand the manifold diameter are deterministically sampled, while those within the diameter range are randomly sampled. Our extensive experiments demonstrate that MaER performs well with tiny buffers and achieves state-of-the-art performance.

\textbf{Limitation.} To calculate the Wasserstein distance, MaER must compute the features of both the teacher and student models. This requires performing inference on the data twice, adding a certain computational burden. In memory management, manifold expansion sampling relies on calculating manifold diameters. Although we simplify the calculation of diameters, the cost of these calculations increases as the number of samples grows. When using large buffers, these computational overheads can make MaER slower to train than other methods.



\bibliography{citation}
\bibliographystyle{ieeetr}

\end{document}